\documentclass[runningheads]{llncs}

\usepackage{eccv}

\usepackage{eccvabbrv}

\usepackage{graphicx}
\usepackage{booktabs}

\usepackage[accsupp]{axessibility}  
\definecolor{darkgreen}{RGB}{0, 100, 0} 
\definecolor{cvprblue}{rgb}{0.21,0.49,0.74}
\usepackage[pagebackref,breaklinks,colorlinks,allcolors=cvprblue]{hyperref}
\usepackage{multirow}
\usepackage[table]{xcolor}  
\usepackage{pifont}
\definecolor{mred}{RGB}{238, 34, 12}
\definecolor{mgreen}{RGB}{1, 127, 0}
\definecolor{mblue}{RGB}{0, 77, 158}
\newcommand{\mredbf}[1]{\textcolor{mred}{\textbf{#1}}}
\newcommand{\mbluebf}[1]{\textcolor{mblue}{\textbf{#1}}}
\usepackage{amsmath}
\usepackage{arydshln} 
\usepackage{makecell}

\usepackage{hyperref}
\newcommand{\MYhref}[3][blue]{\href{#2}{\color{#1}{#3}}} 
\usepackage{orcidlink}
\usepackage{makecell}
\begin{document}
\setlength{\textfloatsep}{2pt plus 0.5pt minus 0.5pt}
\setlength{\dbltextfloatsep}{2pt plus 0.5pt minus 0.5pt}
\setlength{\dblfloatsep}{2pt plus 0.5pt minus 0.5pt}
\setlength{\intextsep}{2pt plus 0.5pt minus 0.5pt}
\setlength{\abovecaptionskip}{2pt plus 0.5pt minus 0.5pt}
\setlength{\belowcaptionskip}{2pt plus 0.5pt minus 0.5pt}
\title{EditHF-1M: A Million-Scale Rich Human Preference Feedback for Image Editing} 

\titlerunning{Abbreviated paper title}

\author{Zitong Xu\inst{1} \and
Huiyu Duan\inst{1*} \and
Zhongpeng Ji\inst{2} \and Xinyun Zhang\inst{3} \and Yutao Liu\inst{4} \and Xiongkuo Min\inst{1} \and Ke Gu\inst{5} \and Jian Zhang\inst{2} \and Shusong Xu\inst{2} \and Jinwei Chen\inst{2} \and Bo Li\inst{2*}\and Guangtao Zhai\inst{1*}\\
*Corresponding Authors}

\authorrunning{Zitong Xu et al.}

\institute{Shanghai Jiao Tong University \and
Vivo Mobile Communication Co., Ltd
\and University of Electronic and Science Technology of China
\and Ocean University of China
\and Beijing University of Technology\\
}
\maketitle

\begin{abstract}
Recent text-guided image editing (TIE) models have achieved remarkable progress, while many edited images still suffer from issues such as artifacts, unexpected editings, unaesthetic contents. Although some benchmarks and methods have been proposed for evaluating edited images, scalable evaluation models are still lacking, which limits the development of human feedback reward models for image editing. To address the challenges, we first introduce \textbf{EditHF-1M}, a million-scale image editing dataset with over 29M human preference pairs and 148K human mean opinion ratings, both evaluated from three dimensions, \textit{i.e.}, visual quality, instruction alignment, and attribute preservation. Based on EditHF-1M, we propose \textbf{EditHF}, a multimodal large language model (MLLM) based evaluation model, to provide human-aligned feedback from image editing. Finally, we introduce \textbf{EditHF-Reward}, which utilizes EditHF as the reward signal to optimize the text-guided image editing models through reinforcement learning. Extensive experiments show that EditHF achieves superior alignment with human preferences and demonstrates strong generalization on other datasets. Furthermore, we fine-tune the Qwen-Image-Edit using EditHF-Reward, achieving significant performance improvements, which demonstrates the ability of EditHF to serve as a reward model to scale-up the image editing. Both the dataset and code will be released in our GitHub repository: \MYhref[magenta]{https://github.com/IntMeGroup/EditHF}{https://github.com/IntMeGroup/EditHF}.
\end{abstract}
\begin{figure}[t]
    \centering
    \includegraphics[width=1\linewidth]{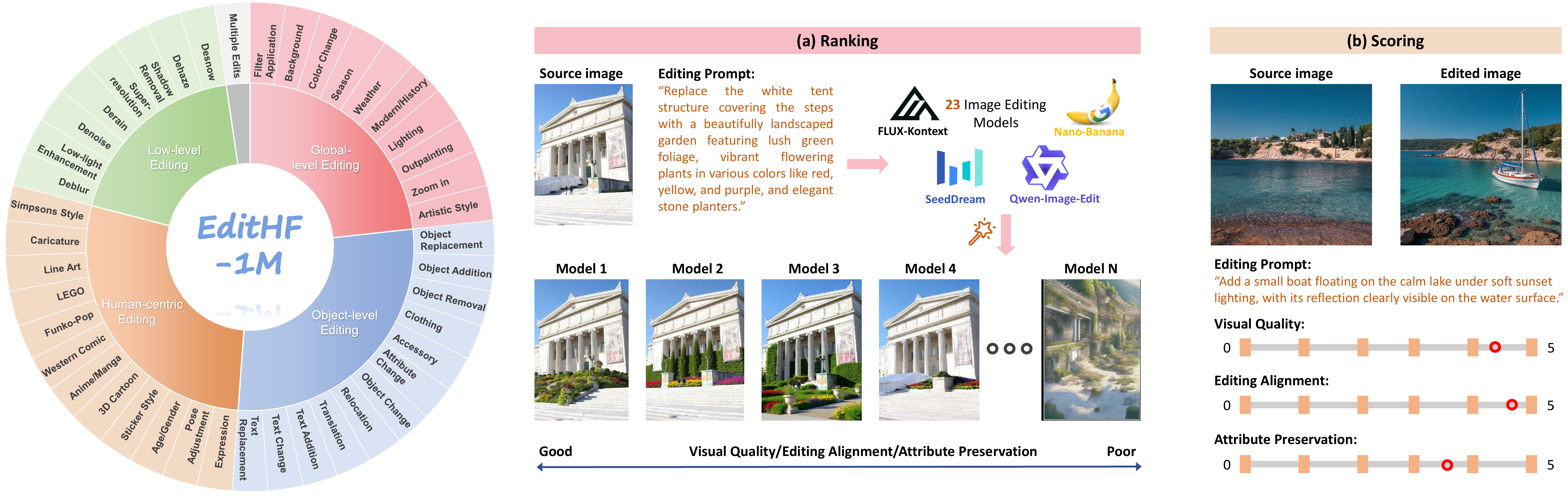}
    \caption{Overview of EditHF-1M and human annotations: (a) Ranking: edited images from different editing models are grouped by source image and editing prompt and ranked to assess relative quality; (b) Scoring: each edited image is rated individually for evaluating absolute quality.}
    \label{bench}
\end{figure}
\begin{table*}[t]
\centering
\caption{Comparison of text-guided image editing model evaluation benchmarks. \textcolor{darkgreen}{\ding{51}} means publicly available, \textcolor{red}{\ding{55}} means not applicable or not available.}
 \resizebox{1\textwidth}{!}{
\begin{tabular}{lcccccccccc}
\toprule
\noalign{\vspace{-1.5pt}}
\multirow{2}{*}{Databases}& Source &Edited & \multirow{2}{*}{Annotations}& Editing & Editing&Prompt& \multirow{2}{*}{Dimensions} & Evaluation&Validated for \\
& Images &Images & & Models &Tasks &Type  & &Model&Refinement \\
\hline
\noalign{\vspace{1.5pt}}
TedBench \cite{TedBench}    & 100&400& 300 Pairs &3 & 4&Description & Overall& \color{red}{\ding{55}}&\color{red}{\ding{55}} \\
EditVal \cite{EditVal}   &648&5,184& 10.3K Scores &8&12&Instruction&Alignment, Preservation&\color{red}{\ding{55}}&\color{red}{\ding{55}}\\
EditEval \cite{ESurvey} & 150&1,050 &4.2K Scores& 7&19&Description, Instruction &Quality, Alignment, Preservation, Realism&\textcolor{darkgreen}{\ding{51}}&\color{red}{\ding{55}}\\
AURORA-Bench \cite{aurora}&400&2,000&
2K Scores
&5&8&Instruction&Overall&\textcolor{darkgreen}{\ding{51}}&\color{red}{\ding{55}} \\
ImagenHub \cite{imagenhub}&179&1,432&
2.8K Scores
&8&-&Description, Instruction&Quality,  Preservation&\color{red}{\ding{55}}&\color{red}{\ding{55}} \\
IE-Bench \cite{IEBench}&301&1,505&
4.5K Scores
&5&11&Description, Instruction&Quality, Correspondence, Preservation&\textcolor{darkgreen}{\ding{51}}&\color{red}{\ding{55}} \\
EBench-18K \cite{lmm4edit}&  1,080&18,360& 
 54K Scores& 17& 21 & Description, Instruction& Quality, Alignment, Preservation&\textcolor{darkgreen}{\ding{51}}& \color{red}{\ding{55}}\\
 EditScore \cite{editscore}&577&2,885& 
 3,072 Pairs& 11& 13 &Instruction& Quality, Alignment, Preservation&\textcolor{darkgreen}{\ding{51}}& \textcolor{darkgreen}{\ding{51}}\\
  EditReward \cite{editreward}&9,557&114.7K& 
200K Pairs& 6& 7 &Instruction& Quality, Alignment&\textcolor{darkgreen}{\ding{51}}& \textcolor{darkgreen}{\ding{51}}\\
\hline
\noalign{\vspace{1.5pt}}
  \rowcolor{gray!20}  
 \textbf{EditHF-1M} &  \textbf{44.3K}&\textbf{1.01M}& \textbf{
29.1M Pairs, 148K Scores}& \textbf{23}& \textbf{43} & \textbf{Description, Instruction}& \textbf{Quality, Alignment, Preservation}&\textcolor{darkgreen}{\ding{51}}&\textcolor{darkgreen}{\ding{51}} \\
 \noalign{\vspace{-1.5pt}}
\bottomrule
\label{comparison}
\end{tabular}}
\end{table*}
\section{Introduction}
The rapid advancement of image editing allows fine-grained image modifications through natural language instructions \cite{nanobanana,seedream4,ACE, qwenedit, ip2p, Magicbrush}. Although some state-of-the-art editing models, such as Nano-Banana \cite{nanobanana}, SeedDream \cite{seedream4}, Qwen-Image-Edit \cite{qwenedit}, \textit{etc.}, have achieved impressive performance, many edited images still suffer from issues such as artifacts, unexpected editings, unaesthetic contents.
Thus, several datasets and benchmarks have emerged for evaluating text-guided image editing (TIE) models \cite{EditVal,ESurvey,I2EBench,IEBench,lmm4edit,editscore,editreward}. However, existing benchmarks exhibit several critical limitations as shown in Table~\ref{comparison}. \textbf{(i) Limited content diversity:} Most existing benchmarks cover only a narrow set of editing tasks and rely on a small number of source images \cite{ESurvey,IEBench,editscore}, which restricts their ability to evaluate flexible and general-purpose image editing models across diverse scenarios. \textbf{(ii) Limited coverage of editing models:} The majority of current datasets include an insufficient number of editing methods \cite{editreward, IEBench} and often exclude recent state-of-the-art or closed-source models \cite{EditVal,ESurvey,I2EBench}, leading to incomplete and potentially biased evaluations that fail to reflect practical model performance. \textbf{(iii) Limited effectiveness for model refinement:} Most existing datasets and benchmarks primarily serve as static evaluation tools and rarely examine or demonstrate the effectiveness for refining image editing models \cite{ESurvey,IEBench,lmm4edit}. 

Besides datasets and benchmarks, some metrics have been developed or applied for evaluating image editing, including image quality assessment (IQA) metrics \cite{NQM, MSSIM, SSIM, FSIM, IFC, VIF, BIQI, BRISQUE, LPIPS}, vision-language approaches \cite{imagereward, clipscore, blip}, and multimodal large language model (MLLM) based evaluation methods \cite{ESurvey, I2EBench}. Traditional IQA metrics primarily focus on natural distortions and low-level visual fidelity, but fail to capture semantic alignment with user instructions. Vision-language approaches such as CLIPScore \cite{clipscore} and HPSv2 \cite{HPS} have shown notable progress in evaluating image generation by incorporating human visual feedback \cite{clipscore,blip,HPS,llavascore,imagereward}, however, they mainly assess text-image alignment and are not designed to capture the semantic changes specific to image editing. More recently, several studies have explored leveraging MLLMs to evaluate image editing \cite{ESurvey, I2EBench, lmm4edit, editscore, editreward}. Nevertheless, zero-shot MLLM-based evaluations \cite{ESurvey,I2EBench} often exhibit limited alignment with human preferences in complex editing scenarios, while fine-tuned MLLMs \cite{lmm4edit,editscore,editreward} tend to suffer from restricted generalization when encountering unseen editing models and editing tasks. Moreover, previous studies have not considered training models by combining score regression and pairwise comparison to achieve both effective evaluation and reward capabilities.

In this work, we introduce \textbf{EditHF-1M}, a million-scale image editing evaluation dataset curated by trained annotators under a rigorous and standardized annotation protocol. EditHF-1M consists of \textbf{44.3K} source images collected from Pico-Banana-400K \cite{picobanana}, EBench-18K \cite{lmm4edit}, and free photography websites, each paired with an editing prompt, covering \textbf{43 editing tasks}, as shown in Figure~\ref{bench}. Based on these source images and prompts, we generate over \textbf{1M} edited images using \textbf{23} state-of-the-art image editing models, including both open-sourced models such as Qwen-Image-Edit \cite{qwenedit} and OmniGen2 \cite{omnigen2}, and closed-source models such as NanoBanana \cite{nanobanana} and Seedream4 \cite{seedream4}. Through an extensive subjective study, we collect over \textbf{29M} human preference pairs and \textbf{148K} human mean opinion ratings, covering perceptual quality, editing alignment, and attribute preservation dimensions. Building upon EditHF-1M, we propose \textbf{EditHF}, an MLLM-based evaluation model for image editing that assesses from multiple dimensions. Extensive experiments conducted on EditHF-1M and other benchmarks demonstrate that EditHF achieves state-of-the-art performance and exhibits strong generalization ability. Furthermore, we introduce \textbf{EditHF-Reward}, which demonstrates that EditHF can be effectively utilized as a reward model for reinforcement learning to enhance image editing models by optimizing Qwen-Image-Edit \cite{qwenedit}.

The main contributions of this work include:
\begin{itemize}
    \item We introduce EditHF-1M, a million scale dataset containing over 1M edited images across diverse tasks with comprehensive human preference rankings and scores covering perceptual quality, editing alignment and attribute preservation dimensions.
    \item We propose EditHF, a unified model by jointly training on score prediction and pairwise comparison, which achieves human-aligned evaluation from multiple dimensions for image editing.
    \item We introduce EditHF-Reward, which utilizes EditHF as the reward signal for reinforcement learning to enhance the ability of image editing models.
    \item Extensive experiments on EditHF-1M and other image editing benchmarks demonstrate that EditHF achieves state-of-the-art performance and generalizes well. Furthermore, EditHF-Reward can effectively enhance the performance of image editing models with EditHF.
\end{itemize}

\section{Related Work}
\subsection{Image Editing}
With the development of generative models such as Stable Diffusion \cite{SD} and FLUX \cite{FLUX}, numerous image editing methods have been proposed \cite{ESurvey}. Based on the type of editing prompts, these methods can be divided into description-based approaches (\textit{e.g.}, “A deer running on the grass” → “A horse running on the grass”) and instruction-based approaches (\textit{e.g.}, “Change the deer to a horse”) \cite{ESurvey, lmm4edit}. Description-based methods mainly use inversion-based modifications \cite{Flowedit, renoise, CDS, PnP, InfEdit}, which first invert an image into a latent representation and then generate edits according to the new prompt. Early instruction-based methods \cite{ip2p, Magicbrush, instructany2pix} train models on instruction-image paired datasets. More recent approaches \cite{qwenedit, omnigen2, dreamomni2, nanobanana, seedream4} unify understanding and generation, interpreting editing instructions and enabling flexible and highly realistic image edits.
\subsection{Benchmarks for Image Editing}
Existing benchmarks for image editing, along with comparisons to our EditHF-1M, are summarized in Table~\ref{comparison}. TedBench \cite{TedBench} is the first image editing benchmark, contains only 100 source images with prompts. EditVal \cite{EditVal} covers more tasks and methods, but the resulting images often suffer from low resolution and blurriness due to data source. EditEval \cite{ESurvey} includes more editing tasks, yet scores derived directly from MLLMs and failed to align with human perception. 
IE-Bench \cite{IEBench} and EBench-18K \cite{lmm4edit} provide Mean Opinion Scores (MOSs), but are limited in image scale. EditScore \cite{editscore} and EditReward \cite{editreward} incorporate pairs of human preferences and demonstrate potential as reward models, but their coverage of editing tasks and source content remains limited, making them less suitable for evaluating flexible editing models. In contrast, EditHF-1M contains over one million edited images spanning 43 editing tasks and 23 editing models, enabling comprehensive evaluation.

\subsection{Evaluation Metrics for Image Editing}
Traditional image quality assessment (IQA) models include full-reference (FR) IQA \cite{SSIM, FSIM, GMSD, LPIPS, STLPIPS, AHIQ} and no-reference (NR) IQA \cite{BRISQUE, BLII, NIQE, CNN, Hyper, MANIQA} metrics. These methods primarily evaluate low-level visual quality and natural distortions, but fail to capture the relationship between the editing instruction and the edited image. To incorporate text prompts, vision-language metrics \cite{imagereward,clipscore,blip,pickscore,llavascore} have been proposed to evaluate alignment for text-based image generation. While effective in measuring image-text correspondence, they often fail to assess the semantic alignment between the editing instruction and the edited image. With the development of MLLMs, many MLLMs demonstrate effectiveness in describing image quality \cite{LMMIQA,harmonyiqa}. Works such as \cite{lmm4edit, editscore, editreward} fine-tune MLLMs on human preference data to provide more accurate assessments. \cite{editscore, editreward} further show that learned evaluation models can serve as reward functions to optimize image editing models via reinforcement learning, bridging the gap between evaluation and generation. However, these learned evaluators typically cover a narrow range of editing tasks and source contents, limiting their applicability to modern, large-scale, and diverse image editing systems. To this end, we propose EditHF to provide human-aligned feedback and EditHF-Reward to utilize EditHF refining image editing models.

\section{EditHF-1M}
In this section, we introduce \textbf{EditHF-1M}, the first million-scale image editing benchmark featuring both human preference pairs and fine-grained scores. The benchmark consists of 44.3K high-quality source images with instruction-based and description-based prompts across 43 editing tasks, over one million edited images generated by 23 image editing models, and large-scale human annotations, including 29.1M preference pairs and 148K scores covering perceptual quality, editing alignment, and attribute preservation. With its diverse image content and extensive annotations, EditHF-1M serves as a comprehensive resource for evaluating image editing models.
\subsection{Design of Editing Tasks} 
Considering both practical relevance and real-world usage, we select 43 image editing tasks, which are categorized into \textbf{global-level tasks}, \textbf{object-level tasks}, \textbf{human-centric tasks} and \textbf{low-level tasks}, as shown in Figure~\ref{bench}. Specifically, global-level tasks involve modifications applied to the entire image, such as overall style transfer or global color adjustment; object-level tasks focus on editing specific objects or regions, such as addition, removal, and attribute modification; human-centric tasks target edits related to human subjects, such as expression change or pose adjustment; and low-level tasks refer to pixel-level transformations aimed at improving image quality, such as denoising, deblurring, and super-resolution. Detailed descriptions of all editing tasks are provided in the supplementary material.
\subsection{Data Collection}
For high-level tasks, approximately 20K source images are collected from publicly available photography websites, each accompanied by a textual description. Based on the image content and the predefined editing task, we leverage the advanced MLLM InternVL3.5 \cite{internvl3_5} to generate corresponding editing prompts, followed by careful manual examination and refinement. To accommodate different types of editing models, the resulting prompts include both instruction-based and description-based formulations. In addition, 26K images are sampled from Pico-Banana-400K \cite{picobanana}, which provides instruction-based editing prompts. For these images, we further employ InternVL3.5 \cite{internvl3_5} to generate complementary description-based prompts conditioned on both the image content and the original instructions, and again conduct manual verification and revision. All the source images in high-level task have a minimum resolution of 1024×1024, which meets or exceeds the maximum input resolution required by most existing image editing models. For low-level tasks, both the source images and the corresponding editing prompts are directly adopted from EBench-18K \cite{lmm4edit}.

Next, we select 23 image editing models, covering both description-based and instruction-based approaches, including both open-source and closed-source methods, with a variety of backbone architectures. Detailed descriptions of these models are provided in the supplementary material. It is noted that some models failed to generate valid edited images for certain source images and prompts, producing abnormal outputs (e.g., noisy or black images). We have filtered these images, but we ensure the images in test set that the images of each editing task and each model are same for further equal comparison. Finally, using our source images and editing prompts, we generate a total of 1.01M edited images.

\subsection{Subjective Experiment}
\label{3.3}
To evaluate the edited images, we conduct a subjective quality assessment experiment based on EditHF-1M, aiming to capture human preferences for edited images and ensure alignment with real-world human perception. Unlike existing benchmarks, which typically annotate scores or rankings alone, we collect both group rankings and absolute scores as shown in Figure~\ref{bench}. 

Group rankings are particularly effective for comparisons among results produced by different models under the same source image and editing prompt, as they reduce individual scoring bias and provide more reliable relative judgments. This makes rankings well suited for model-level comparison in specific editing cases. However, rankings are limited to local comparisons and cannot be directly aggregated across different editing tasks or image contents. In contrast, absolute scoring enables cross-task and cross-content comparisons, allowing us to analyze model performance across diverse editing scenarios and to study strengths and weaknesses with respect to specific editing tasks. By jointly leveraging rankings and scores, our evaluation protocol combines the complementary advantages of both, enabling reliable assessment of image editing models.
\begin{figure}[t]
    \centering
    \includegraphics[width=1\linewidth]{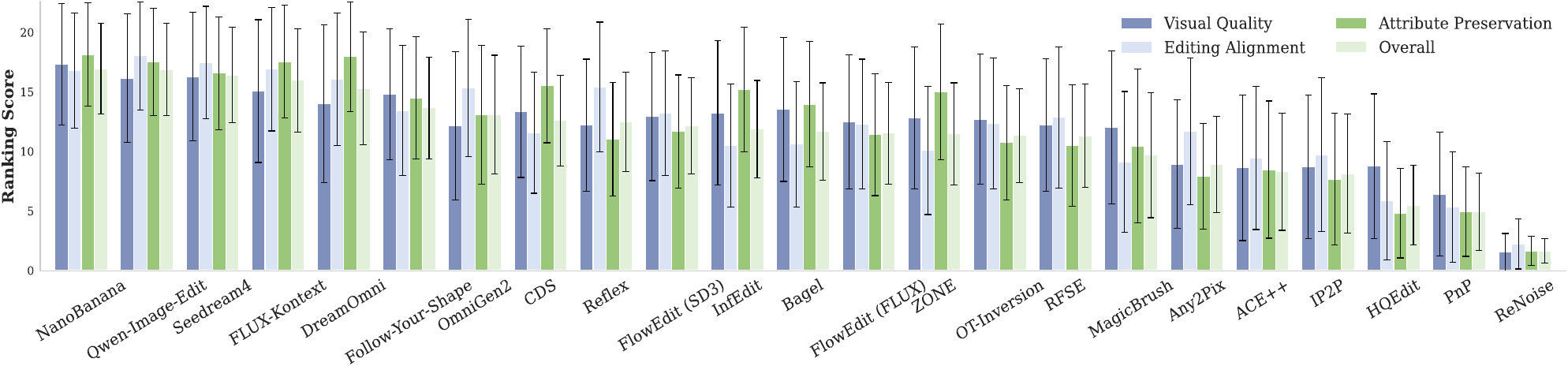}
    \caption{Comparison of image editing models using EditHF-1M. Ranking scores are derived from win counts in pairwise group comparisons.}
    \label{modelcomparison}
\end{figure}
To comprehensively evaluate image editing, we perform the subjective evaluation from three dimensions:
\begin{itemize}
   \item \textbf{Perceptual Quality:} measures the overall visual quality of the edited image, such as realism, absence of artifacts, structural integrity, color consistency, and the richness of visual details.

\item \textbf{Editing Alignment:} evaluates the degree to which the edited image correctly follows the given editing instruction, assessing whether the intended modifications are accurately and completely applied.

\item \textbf{Attribute Preservation:} assesses how well the edited image maintains essential attributes of the source image beyond the intended edits, including preservation of subject identity, key visual characteristics, and contextual consistency. 
\end{itemize}  
The experiment is conducted using a Python-based graphical user interface displayed on a calibrated LED monitor with a resolution of 3840 × 2160, where images are presented at a resolution of 1024 × 1024. Annotators are seated approximately 2 feet from the monitor in a controlled environment. Each image ranking group is annotated by three professional annotators, while each score is obtained from ratings provided by five professional annotators to ensure reliability. To further ensure the reliability of the test set, its scores are obtained from ratings provided by fifteen professional annotators. All annotators undergo rigorous training following a standardized protocol. A pre-test is conducted to assess participants’ comprehension of the criteria and their alignment with the standard examples. Participants who did not meet the required accuracy threshold were excluded from further participation.


For ranking annotation, EditHF-1M is divided into 44.3K image groups, where each group contains edited images produced by different editing models given the same source image and editing instruction. Annotators are asked to rank the edited images along the three evaluation dimensions independently. Rankings are collected from three annotators per group, and inter-annotator consistency is assessed. Groups exhibiting low agreement are re-annotated by additional three annotators to ensure reliable and consistent rankings. The final ranking for each image group is computed as the average rank from the three annotators. We further generate human preference pairs based on the rankings, for the group including $M$ images, $C_M^2$ preference pairs generated.

For scoring annotation, 49.4K edited images from EditHF-1M across all the editing tasks and models are selected, and these samples are ensured with diverse source image content and editing prompts. Participants are randomly shown a source image, an edited image, and the corresponding editing prompt, and asked to rate the edited image on a 5-point continuous scale across three aspects. Following \cite{subject}, those beyond 2 standard deviations from the mean are regarded as outliers and excluded, and participants with over 5\% outlier ratings are removed. Remaining ratings are converted to Z-scores and linearly scaled to [0,100]. The final score for each image is computed as follows:
\begin{equation}
    z_{ij} = \frac{s_{ij} - \mu_i}{\sigma_i}, \ z_j = \frac{1}{N_j} \sum_{i=1}^{N_j} z_{ij}, \
    \text{Score}_j = \frac{100(z_j + 3)}{6}
\end{equation}
where $s_{ij}$ is the raw score given by the i-th subject to the j-th image, $\mu_i$ is the mean rating and $\sigma_i$ is the standard deviation provided by the i-th subject and $N_j$ is the number of ratings for the j-th image. Scores and pairs are further compared and any annotations where the scores and preference pairs are inconsistent are rejected. Detailed descriptions of the annotation protocol and procedures are provided in Supplementary Material, Section 4. in Finally, we have obtained 29.1M pairs and 148K scores from three evaluation dimensions.

\subsection{Data Analysis}
\begin{figure}[t]
    \centering
    \includegraphics[width=1\linewidth]{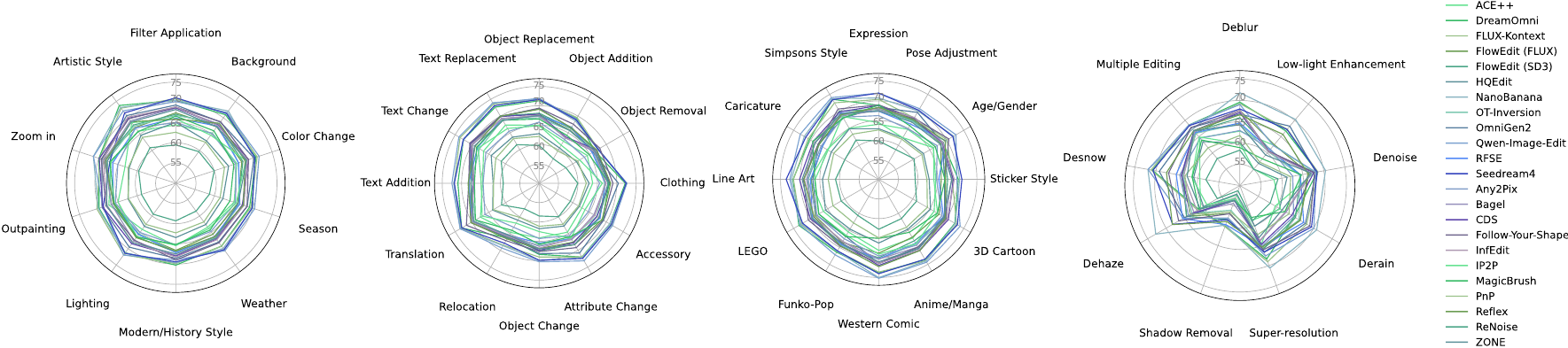}
    \caption{Comparison of image editing models across different editing tasks.}
    \label{catcomparison}
\end{figure}

To benchmark the editing models, we use group rankings from EditHF-1M. Each image is assigned a ranking score based on its win count across these pairwise comparisons. An overall score is then derived by combining the visual quality score $\text{Score}_{v}$, the editing alignment score $\text{Score}_{e}$, and the attribute preservation score $\text{Score}_{p}$ according to the following equation:
\begin{equation}
\text{Score}_{all}=\text{Score}_{v}^{0.3}\times\text{Score}_{e}^{0.4}\times\text{Score}_{p}^{0.3}
\label{overall_score}
\end{equation}
The editing alignment score is assigned a higher weight to emphasize the importance of satisfying editing instructions. Figure~\ref{modelcomparison} shows the results of the editing models in EditHF-1M. Qwen-Image-Edit \cite{qwen3} performs best in editing alignment while NanoBanana \cite{nanobanana} perform best in other evaluation dimensions.

We further benchmark each model’s performance across different editing tasks. Since group rankings cannot reflect the absolute quality of individual edits, we utilize the parts of EditHF-1M that provide explicit score annotations. For each task, a task-level score is computed by averaging the scores over all editing samples within that task. As shown in Figure~\ref{catcomparison}, models generally perform better on high-level editing tasks, indicating strong capabilities in handling semantic-level modifications. Among the high-level tasks, pose adjustment and object relocation exhibit lower performance, suggesting that precise geometric control and spatial consistency remain challenging. Low-level editing tasks show overall weaker results, reflecting limitations in fine-grained pixel-level editing. Shadow removal performs the worst among all tasks, highlighting the difficulty of modeling complex illumination interactions and preserving visual realism.
\section{EditHF}
In this section, we present EditHF, a unified image editing evaluation model that supports both fine-grained quality score prediction and pairwise preference comparison, jointly assessing perceptual quality, editing alignment, and attribute preservation in alignment with human perception.

\subsection{Model Architecture}
\begin{figure}[t]
    \centering
    \includegraphics[width=1\linewidth]{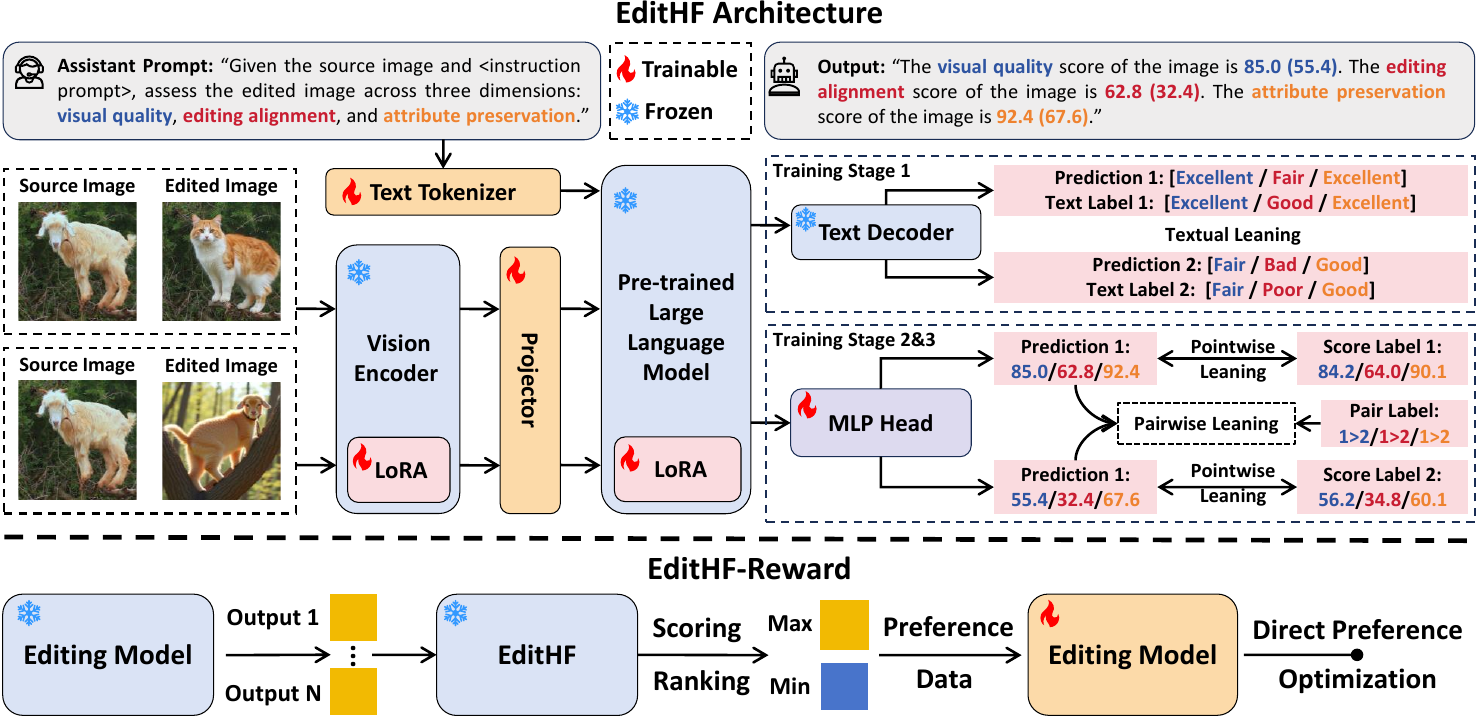}
    \caption{Overview of the EditHF architecture and the editing model refinement process guided by EditHF.}
    \label{model}
\end{figure}

Inspired by the success of MLLMs in multimodal reasoning, we adopt an MLLM as the backbone for EditHF. 
The model jointly reasons over an edited image $I_e$, the corresponding source image $I_s$, and the editing prompt $T$, producing fine-grained scores on three dimensions, as shown in Figure~\ref{model}.

Visual features are first extracted from $I_e$ and $I_s$ using a pre-trained vision encoder $E_v$ based on InternViT \cite{internViT}. 
These features are then projected into the language embedding space via a trainable projector $P_\varphi$ with parameters $\varphi$, yielding the visual tokens
\begin{equation}
T_e = P_\varphi\!\left(E_v(I_e)\right), \qquad
T_s = P_\varphi\!\left(E_v(I_s)\right).
\end{equation}
Meanwhile, the editing prompt is tokenized into text tokens $T_p$. The visual and textual tokens are concatenated and fed into an advanced MLLM, InternVL3.5 \cite{internvl3_5}, to perform multimodal feature fusion and joint reasoning. Defining the LLM backbone as $H_{\psi}$ with trainable parameters $\psi$, the last hidden states are obtained as $\mathbf{h} = H_{\psi}\!\left(T_e, T_s, T_p\right)$.
To support both textual response generation and fine-grained score prediction, EditHF employs an adaptive decoding strategy. The last hidden states $\mathbf{h}$ are first decoded by a text decoder, enabling the model to acquire an initial understanding and generate human-interpretable responses. Once the text output stabilizes, the same hidden states are fed into an MLP head,  $S_{\omega}$ with parameters $\omega$, to produce quality scores $s_v, s_e, s_p$ corresponding to the three dimensions. The whole process can be formally written as:
\begin{equation}
s_{v}, s_{e}, s_{p} = S_\omega(H_\psi(P_\varphi(E_v(I_s,I_e)),T_p)),
\end{equation}
where the parameters $\varphi, \psi, \omega$ are optimized during training. The output scores are used to assess perceptual quality, editing alignment, and attribute preservation, and can be further employed for pairwise preference comparison. 
\subsection{Training Strategy}
Our training consists of three stages: \textbf{textual learning, pointwise learning, and pairwise learning}, each with a specific loss function targeting distinct objectives. 

\textbf{Textual learning} serves as the initial stage, providing the model with a coarse-grained understanding of image editing quality through textual categories. Continuous quality scores are converted into discrete, text-based quality levels, leveraging the fact that MLLMs often exhibit stronger comprehension of textual information than raw numerical values. Specifically, we uniformly divide the range between the maximum score \( M \) and the minimum score \( m \) into five distinct intervals, and assign each score \( s \) to a corresponding quality level \( L(s) \):
\begin{equation}
L(s) = l_i \quad \text{if} \quad m + \frac{i-1}{5}(M - m) < s \leq m + \frac{i}{5}(M - m)
\end{equation}
where \( i \in \{1, 2, 3, 4, 5\} \) corresponds to the five standard quality levels, \( \{l_i\}_{i=1}^5 = \{\text{bad}, \text{poor}, \text{fair}, \text{good}, \text{excellent}\} \) are the standard five-point rating levels. This stage is trained using a \textbf{cross-entropy loss}.

\textbf{Pointwise learning} aims to provide the model the ability to predict fine-grained quality scores. Leveraging samples with human-annotated scores from EditHF-1M, the model learns to estimate absolute quality values, reflecting direct perceptual judgments. This stage is supervised using the \textbf{mean squared error loss}, which encourages the model to approximate human-provided scores as closely as possible.

\textbf{Pairwise learning} focuses on relative quality assessment among images sharing the same source and editing instructions. Preference pairs derived from group rankings in EditHF-1M are used to train the model. Pairs with larger rank differences are prioritized during training, followed by pairs with smaller differences, facilitating a coarse-to-fine understanding of relative quality. The model is optimized with the following \textbf{pairwise loss}:
\begin{equation}
\mathcal{L}_{\text{pairwise}} = \frac{1}{N} \sum_{i=1}^{N} \log\Big( 1 + \exp(s_{\text{neg},i} - s_{\text{pos},i}) \Big),
\end{equation}
where $s_{\text{pos},i}$ and $s_{\text{neg},i}$ denote the predicted scores for the preferred and less-preferred images in the i-th pair. This loss encourages the model to assign higher scores to perceptually superior images, enhancing its sensitivity to relative image editing quality.

While pointwise training enables the model to predict absolute quality scores that reflect direct human perception, it is inherently insensitive to subtle relative differences among images derived from the same source and editing instruction. Conversely, pairwise training focuses on modeling relative preferences within groups, capturing fine-grained distinctions between similar edits, but it lacks the ability to produce absolute scores that are comparable across different sources or editing instructions. By combining both training paradigms, our model acquires both the capability of absolute score prediction and relative quality comparison. This synergistic training strategy ensures that the model can serve as a reliable evaluation benchmark, while also functioning as a reward model for preference-guided optimization in image editing.
\begin{table*}[t]

\caption{
Performance comparisons of quality evaluation methods on EditHF-1M from perspectives of \underline{Visual Quality}, \underline{Editing Alignment}, and \underline{Attribute Preservation}. 
$\mathrm{SRCC}_{global}$ and $\mathrm{PLCC}_{global}$ are reported to measure correlation over the entire dataset. 
$\mathrm{SRCC}_{group}$ denotes the average SRCC computed across all image groups.
Acc represents the accuracy of pairwise comparison prediction. 
$\spadesuit$ Traditional FR IQA metrics, $\heartsuit$ traditional NR IQA metrics, $\clubsuit$ deep learning-based FR IQA methods, $\diamondsuit$ deep learning-based NR IQA methods, \ding{72} vision-language methods, \ding{73} MLLM-based models. 
The fine-tuned results are marked with \raisebox{0.5ex}{\scriptsize \ding{91}}. 
The best results are highlighted in \mredbf{red}, and the second-best results are highlighted in \mbluebf{blue}.
}
\centering
\resizebox{1\textwidth}{!}{\begin{tabular}{l||cccc|cccc|cccc}
\toprule
Dimension& \multicolumn{4}{c}{Visual Quality} & \multicolumn{4}{c}{Editing Alignment} & \multicolumn{4}{c}{Attribute Preservation} \\
\cmidrule(lr){2-5}
\cmidrule(lr){6-9}
\cmidrule(lr){10-13}
Methods/Metrics & $\mathrm{SRCC}_{global}$ & $\mathrm{PLCC}_{global}$ & $\mathrm{SRCC}_{group}$ & Acc & $\mathrm{SRCC}_{global}$ & $\mathrm{PLCC}_{global}$ & $\mathrm{SRCC}_{group}$ & Acc& $\mathrm{SRCC}_{global}$ & $\mathrm{PLCC}_{global}$ & $\mathrm{SRCC}_{group}$ & Acc \\
\hline
$\heartsuit$DIIVINE\cite{DIIVINE}  & 0.2062 & 0.3162 & 0.1795 & 0.5633 & 0.0618 & 0.1947 & 0.0962 & 0.5332 & 0.1641 & 0.2686 & 0.1506 & 0.5522 \\
$\heartsuit$BRISQUE\cite{BRISQUE} & 0.2446 & 0.2996 & 0.2020 & 0.4286 & 0.0494 & 0.0975 & 0.0495 & 0.4827 & 0.2019 & 0.2457 & 0.1654 & 0.4426 \\
$\heartsuit$NIQE\cite{NIQE}  & 0.2816 & 0.3145 & 0.2386 & 0.4142 & 0.0635 & 0.1432 & 0.0811 & 0.4723 & 0.2623 & 0.2778 & 0.2405 & 0.4157 \\
\hdashline
$\spadesuit$MSE & 0.1421 & 0.1418 & 0.2900 & 0.3968 & 0.0600 & 0.0601 & 0.1730 & 0.4412 & 0.3108 & 0.1893 & 0.4736 & 0.3268 \\
$\spadesuit$PSNR & 0.1270 & 0.0232 & 0.2844 & 0.6011 & 0.0572 & 0.0134 & 0.1694 & 0.5575 & 0.2978 & 0.1851 & 0.4690 & 0.6708 \\
$\spadesuit$GMSD \cite{GMSD} & 0.0176 & 0.0222 & 0.2715 & 0.4019 & 0.0527 & 0.0486 & 0.1903 & 0.4339 & 0.1926 & 0.1398 & 0.4750 & 0.3265 \\
$\spadesuit$MSSIM \cite{MSSIM} & 0.1023 & 0.1041 & 0.2360 & 0.5821 & 0.0894 & 0.0915 & 0.1768 & 0.5597 & 0.2914 & 0.2490 & 0.4563 & 0.6628 \\
$\spadesuit$SSIM \cite{SSIM} & 0.0807 & 0.0762 & 0.2454 & 0.5868 & 0.0688 & 0.0646 & 0.1641 & 0.5557 & 0.2448 & 0.2107 & 0.4480 & 0.6605 \\
$\spadesuit$SCSSIM \cite{SCSSIM} & 0.0398 & 0.0226 & 0.0867 & 0.5283 & 0.0337 & 0.0329 & 0.0567 & 0.5187 & 0.1140 & 0.1054 & 0.3029 & 0.6053 \\
$\spadesuit$VIF \cite{VIF} & 0.1769 & 0.0134 & 0.3703 & 0.6343 & 0.2009 & 0.0922 & 0.3217 & 0.6133 & 0.3367 & 0.1167 & 0.5625 & 0.7087 \\
$\spadesuit$FSIM \cite{FSIM} & 0.1777 & 0.1840 & 0.3357 & 0.6219 & 0.1350 & 0.1461 & 0.2461 & 0.5846 & 0.3683 & 0.3271 & 0.5387 & 0.6993 \\
\hdashline
$\diamondsuit$CNNIQA\raisebox{0.5ex}{\scriptsize \ding{91}} \cite{CNN} & 0.6438 & 0.5824 & 0.6394 & 0.7609 & 0.1927 & 0.2066 & 0.1719 & 0.5605 & 0.3229 & 0.2153 & 0.3086 & 0.6149 \\
$\diamondsuit$MANIQA\raisebox{0.5ex}{\scriptsize \ding{91}} \cite{MANIQA} & 0.8150 & 0.8104 & 0.7893 & 0.8152 & 0.3068 & 0.3381 & 0.2569 & 0.5907 & 0.5090 & 0.5192 & 0.4968 & 0.6853 \\
$\diamondsuit$TOPIQ\raisebox{0.5ex}{\scriptsize \ding{91}} \cite{TOPIQ} & 0.7926 & 0.8007 & 0.7359 & 0.7823 & 0.2977 & 0.3175 & 0.2630 & 0.5926 & 0.6006 & 0.6403 & 0.5314 & 0.6936 \\
$\diamondsuit$Q-align\raisebox{0.5ex}{\scriptsize \ding{91}} \cite{qalign} & \mbluebf{0.8285} & \mbluebf{0.8403} & 0.7659 & 0.7948 & 0.3322 & 0.3520 & 0.3340 & 0.6189 & 0.6401 & 0.7023 & 0.5639 & 0.7049 \\
\hdashline
$\clubsuit$LPIPS \cite{LPIPS} & 0.2760 & 0.2629 & 0.3664 & 0.3666 & 0.1783 & 0.1690 & 0.2920 & 0.3988 & 0.4783 & 0.4192 & 0.6007 & 0.2758 \\
$\clubsuit$ST-LPIPS \cite{STLPIPS} & 0.2896 & 0.2953 & 0.3763 & 0.3611 & 0.1825 & 0.1986 & 0.2903 & 0.3992 & 0.4919 & 0.4376 & 0.6082 & 0.2710 \\
$\clubsuit$CVRKD\raisebox{0.5ex}{\scriptsize \ding{91}} \cite{CVRKD} & 0.7649 & 0.7777 & 0.7430 & 0.7917 & 0.3282 & 0.3344 & 0.2943 & 0.6039 & 0.7831 & 0.7989 & 0.7931 & 0.8155 \\
$\clubsuit$AHIQ\raisebox{0.5ex}{\scriptsize \ding{91}} \cite{AHIQ} & 0.8172 & 0.8249 & 0.8219 & 0.8326 & 0.3538 & 0.3680 & 0.3254 & 0.6157 & 0.8110 & 0.8210 & 0.8223 & 0.8308 \\
\hdashline
\ding{72}CLIPScore \cite{clipscore} & 0.0348 & 0.0487 & 0.0879 & 0.4659 & 0.1430 & 0.1585 & 0.1085 & 0.5377 & 0.0422 & 0.0569 & 0.1035 & 0.4597 \\
\ding{72}BLIPScore \cite{blip} & 0.1158 & 0.1235 & 0.0171 & 0.4921 & 0.2562 & 0.2580 & 0.2185 & 0.5805 & 0.1330 & 0.1376 & 0.0345 & 0.4861 \\
\ding{72}ImageReward \cite{imagereward} & 0.0973 & 0.1019 & 0.0555 & 0.5188 & 0.2017 & 0.1894 & 0.2340 & 0.5853 & 0.1092 & 0.1195 & 0.0269 & 0.5062 \\
\ding{72}HPSv2 \cite{HPS} & 0.1128 & 0.1291 & 0.0280 & 0.5098 & 0.2912 & 0.3216 & 0.2429 & 0.5851 & 0.2278 & 0.2484 & 0.1643 & 0.5571 \\
\ding{72}VQAScore \cite{vqa} & 0.0904 & 0.0930 & 0.0029 & 0.4999 & 0.2763 & 0.2850 & 0.2536 & 0.5900 & 0.3573 & 0.3754 & 0.2910 & 0.6032 \\
\hdashline
\ding{73}LLaVA-1.5 (7B) \cite{llava} & 0.0700 & 0.0942 & 0.0729 & 0.1677 & 0.0253 & 0.0271 & 0.0163 & 0.0396 & 0.1463 & 0.1548 & 0.1122 & 0.1100 \\
\ding{73}mPLUG-Owl3 (7B) \cite{mplug} & 0.1435 & 0.0801 & 0.1242 & 0.5338 & 0.1173 & 0.0645 & 0.1601 & 0.5606 & 0.1962 & 0.0331 & 0.2368 & 0.5932 \\
\ding{73}MiniCPM-V2.6 (8B) \cite{minicpm} & 0.1482 & 0.0913 & 0.1224 & 0.3648 & 0.2283 & 0.0220 & 0.2473 & 0.4098 & 0.2765 & 0.0671 & 0.2479 & 0.4029 \\
\ding{73}Ovis2.5 (8B) \cite{ovis25} & 0.2011 & 0.0984 & 0.1858 & 0.4255 & 0.2274 & 0.0377 & 0.3004 & 0.4647 & 0.2467 & 0.0716 & 0.2385 & 0.4376 \\
\ding{73}LLama3.2-V (11B) \cite{llama} & 0.4004 & 0.2283 & 0.3115 & 0.2813 & 0.3743 & 0.3334 & 0.3337 & 0.2071 & 0.2181 & 0.1944 & 0.0945 & 0.1526 \\
\ding{73}DeepSeekVL2 (small) \cite{deepseekv2} & 0.1634 & 0.1081 & 0.0824 & 0.3440 & 0.1751 & 0.0663 & 0.1491 & 0.3335 & 0.2562 & 0.1232 & 0.1781 & 0.3408 \\
\ding{73}Qwen2.5-VL (8B) \cite{qwenvl2}& 0.2818 & 0.3065 & 0.3362 & 0.4033 & 0.4528 & 0.4572 & 0.6193 & 0.5195 & 0.4664 & 0.4278 & 0.5871 & 0.4869 \\
\ding{73}Qwen3-VL (8B) \cite{qwen3}& 0.3256 & 0.2263 & 0.2982 & 0.5096 & 0.4578 & 0.4606 & 0.6237 & 0.4990 & 0.4791 & 0.4415 & 0.6296 & 0.4884 \\
\ding{73}InternVL3 (8B) \cite{internvl3}& 0.2462 & 0.2995 & 0.2947 & 0.4710 & 0.4649 & 0.4354 & 0.5853 & 0.5851 & 0.3604 & 0.3095 & 0.4335 & 0.5512 \\
\ding{73}InternVL3.5 (8B) \cite{internvl3_5} & 0.3607 & 0.3055 & 0.3976 & 0.3148 & 0.4671 & 0.3700 & 0.6051 & 0.4103 & 0.5065 & 0.3324 & 0.5887 & 0.4077 \\
\ding{73}EditScore (Qwen3) \cite{editscore} & 0.3678 & 0.3297 & 0.3730 & 0.6440 & 0.3700 & 0.2657 & 0.4149 & 0.6753 & 0.4110 & 0.3054 & 0.4639 & 0.6863 \\
\ding{73}EditReward (MiMo) \cite{editreward}& 0.3940 & 0.3320 & 0.4321 & 0.3412 & 0.3410 & 0.3290 & 0.3705 & 0.3657 & 0.3941 & 0.3544 & 0.4266 & 0.3449 \\
\ding{73}DeepSeekVL2 (small)\raisebox{0.5ex}{\scriptsize \ding{91}} \cite{deepseekv2} & 0.7931 & 0.7771 & 0.8111 & 0.8330 & 0.8176 & 0.8077 & 0.8273 & 0.8341 & 0.7790 & 0.7984 & 0.8870 & 0.8693 \\
\ding{73}Qwen3-VL (8B)\raisebox{0.5ex}{\scriptsize \ding{91}} \cite{qwen3} & 0.7751 & 0.7673 & \mbluebf{0.8674} & 0.8588 & 0.8246 & 0.8137 & 0.8498 & 0.8472 & \mbluebf{0.8173} & \mbluebf{0.8342} & \mbluebf{0.9167} & \mbluebf{0.8904} \\
\ding{73}InternVL3 (8B)\raisebox{0.5ex}{\scriptsize \ding{91}} \cite{internvl3} & 0.8044 & 0.7852 & 0.8284 & \mbluebf{0.8817} & \mbluebf{0.8248} & \mbluebf{0.8156} & \mbluebf{0.8540} & \mbluebf{0.8498} & 0.8120 & 0.8297 & 0.9137 & 0.8883 \\
\hline
  \rowcolor{gray!20}  
EditHF (Ours)\raisebox{0.5ex}{\scriptsize \ding{91}} & \mredbf{0.8548} & \mredbf{0.8422} & \mredbf{0.9223} & \mredbf{0.9152} & \mredbf{0.8488} & \mredbf{0.8348} & \mredbf{0.9319} & \mredbf{0.9298} & \mredbf{0.8475} & \mredbf{0.8442} & \mredbf{0.9251} & \mredbf{0.9197} \\
\noalign{\vspace{-1.5pt}}
\bottomrule
\end{tabular}}
\label{performances}
\end{table*}

\section{Experiments}
In this section, we evaluate the performance of our EditHF through extensive experiments.
\subsection{Experiment Setup}
We train the EditHF on our EditHF-1M, which is split to training, validating and testing set with a 5:1:1 ratio. Each image in the testing set includes both ranking and score annotation. We utilize the LoRA \cite{lora} for fine-tuning, with LoRA modules are applied to both vision model and language model. The models are implemented with PyTorch and trained on 2 NVIDIA RTX A6000 GPU with a batch size of 16. 
\subsection{Evaluation on EditHF-1M}
Table~\ref{performances} presents the performance of EditHF compared with traditional and deep learning-based FR IQA and NR IQA methods, vision-language models, and MLLM-based approaches. We report the Spearman Rank Correlation Coefficient (SRCC) and Pearson Linear Correlation Coefficient (PLCC) over the entire test set to evaluate overall score prediction, denoted as $\mathrm{SRCC}_{global}$ and $\mathrm{PLCC}_{global}$. To evaluate pairwise comparisons, we report the average SRCC and comparison accuracy among image groups sharing the same source image and editing prompt, denoted as $\mathrm{SRCC}_{group}$ and Acc. 

Traditional IQA methods perform poorly, as their features are primarily designed to capture low-level distortions and are ineffective for high-level semantic changes. Deep learning-based IQA methods achieve better results in visual quality assessment but still struggle to evaluate editing alignment. Vision-language pretraining models show limited effectiveness because they focus on text-image alignment and fail to capture editing semantics. Zero-shot performance of MLLM-based methods is also poor due to insufficient understanding of the editing evaluation task. EditScore \cite{editscore} and EditReward \cite{editreward}, which are trained on other image editing datasets, still exhibit limited performance because their training data covers only a restricted range of image content, editing tasks, and editing models, leading to insufficient generalization ability. Our EditHF achieves the best performance in aligning with human perception across all evaluation dimensions. Table~\ref{performances} further shows that MLLMs fine-tuned on our EditHF-1M achieve significant improvements, demonstrating the value of our dataset for advancing image editing quality assessment.

Table~\ref{compare_TIE} demonstrates the effectiveness of our model to benchmark image editing models. Our model achieves the highest SRCC with human scores, demonstrating its strong capability in benchmarking image editing models. Moreover, the results reveal a clear performance gap between closed-source image editing methods, such as NanoBanana~\cite{nanobanana}, Seedream4~\cite{seedream4}, and FLUX-Kontext~\cite{fluxkontext}, and open-source models, motivating us to further refine open-source models using data derived from closed-source methods.
\begin{table*}[t]
\belowrulesep=0pt
\aboverulesep=0pt
\centering
\renewcommand{\arraystretch}{0.85}
\caption{Comparisons of the alignment between different evaluation methods and human perception in evaluating image editing models. The best results are highlighted in \mredbf{red}, and the second-best results are highlighted in \mbluebf{blue}. \raisebox{0.5ex}{\scriptsize \ding{91}} denotes fine-tuned models.} 
\resizebox{\textwidth}{!}{
\begin{tabular}{l||c:ccc| c:ccc| c:ccc| c:ccc| c:c}
\toprule
\noalign{\vspace{1.5pt}}
Dimensions & \multicolumn{4}{c}{Perceptual Quality} & \multicolumn{4}{c}{Editing Alignment} & \multicolumn{4}{c}{Attribute Preservation}& \multicolumn{4}{c}{Overall Score}& \multicolumn{2}{c}{Overall Rank}\\
\cmidrule(lr){2-5}
\cmidrule(lr){6-9}
\cmidrule(lr){10-13}
\cmidrule(lr){14-17}
\cmidrule(lr){18-19}

\noalign{\vspace{1.5pt}}
Models/Metrics & Human &\cellcolor{gray!20}Ours\raisebox{0.5ex}{\scriptsize \ding{91}} & Q-align\raisebox{0.5ex}{\scriptsize \ding{91}} & EditReward & Human &\cellcolor{gray!20}Ours\raisebox{0.5ex}{\scriptsize \ding{91}}  &Qwen3-VL\raisebox{0.5ex}{\scriptsize \ding{91}}&EditReward &Human&\cellcolor{gray!20}Ours\raisebox{0.5ex}{\scriptsize \ding{91}}&AHIQ\raisebox{0.5ex}{\scriptsize \ding{91}}&EditScore&Human&\cellcolor{gray!20}Ours\raisebox{0.5ex}{\scriptsize \ding{91}} & EditScore&EditReward&Human&\cellcolor{gray!20}Ours\\

\hline
\noalign{\vspace{1.5pt}}
NanoBanana \cite{nanobanana} & 17.54 & \cellcolor{gray!20}17.61 & 16.20 & 5.863 & 17.01 &\cellcolor{gray!20}17.97 & 17.24 & 6.746 & 18.56 & \cellcolor{gray!20}18.43 & 18.65 & 18.85 & 17.68 & \cellcolor{gray!20}17.70 & 5.608 & 1.810 & 1 & \cellcolor{gray!20}1 \\
Qwen-Image-Edit \cite{qwenedit} & 15.26 & \cellcolor{gray!20}16.53 & 14.51 & 6.453 & 18.17 & \cellcolor{gray!20}18.26 & 17.38 & 6.952 & 17.51 & \cellcolor{gray!20}17.86 & 19.95 & 18.19 & 17.15 & \cellcolor{gray!20}17.33 & 5.463 & 1.799 & 2 & \cellcolor{gray!20}2 \\
Seedream4 \cite{seedream4} & 15.94 &\cellcolor{gray!20}16.77 & 14.85 & 6.377 & 17.31 & \cellcolor{gray!20}18.06 & 17.31 & 7.013 & 16.57 & \cellcolor{gray!20}17.38 & 19.34 & 17.85 & 16.74 & \cellcolor{gray!20}17.19 & 5.352 & 1.799 & 3 & \cellcolor{gray!20}4 \\
FLUX-Kontext \cite{fluxkontext} & 15.12 & \cellcolor{gray!20}16.37 & 14.87 & 6.602 & 17.12 & \cellcolor{gray!20}18.04 & 17.13 & 7.126 & 17.40 & \cellcolor{gray!20}18.06 & 20.13 & 17.05 & 16.65 & \cellcolor{gray!20}17.25 & 5.259 & 1.794 & 4 & \cellcolor{gray!20}3 \\
DreamOmni \cite{dreamomni2} & 14.33 & \cellcolor{gray!20}15.91 & 13.57 & 6.658 & 16.14 & \cellcolor{gray!20}17.45 & 16.69 & 7.109 & 18.08 & \cellcolor{gray!20}18.22 & 20.06 & 16.59 & 16.19 & \cellcolor{gray!20}16.93 & 5.234 & 1.794 & 5& \cellcolor{gray!20}5 \\
Follow-Your-Shape \cite{followyourshape} & 14.85 & \cellcolor{gray!20}16.41 & 14.81 & 10.29 & 12.94 &\cellcolor{gray!20}16.53 & 14.81 & 11.01 & 14.24 & \cellcolor{gray!20}15.80 & 16.20 & 13.78 & 13.92 & \cellcolor{gray!20}16.01 & 4.664 & 1.695 & 6 & \cellcolor{gray!20}6 \\
OmniGen2 \cite{omni2} & 11.89 &\cellcolor{gray!20}15.16 & 12.60 & 10.21 & 15.25 & \cellcolor{gray!20}16.86 & 16.33 & 10.32 & 13.26 & \cellcolor{gray!20}15.67 & 18.39& 13.41 & 13.66 & \cellcolor{gray!20}15.73 & 4.581 & 1.704 & 7 & \cellcolor{gray!20}7 \\
Reflex \cite{reflex} & 12.03 & \cellcolor{gray!20}15.47 & 13.03 & 10.87 & 14.83 & \cellcolor{gray!20}16.59 & 15.89 & 11.08 & 11.11 & \cellcolor{gray!20}14.48 & 18.07 & 11.99 & 12.89 & \cellcolor{gray!20}15.36 & 4.413 & 1.686 & 8 & \cellcolor{gray!20}8 \\
CDS \cite{CDS} & 13.11 & \cellcolor{gray!20}15.59 & 13.29 & 11.57 & 11.07 & \cellcolor{gray!20}14.71 & 14.15 & 11.95 & 14.65 & \cellcolor{gray!20}16.50 & 12.47 & 10.59 & 12.71 & \cellcolor{gray!20}15.23 & 4.370 & 1.665 & 9& \cellcolor{gray!20}9 \\
InfEdit \cite{InfEdit} & 13.10 & \cellcolor{gray!20}15.49 & 13.52 & 10.98 & 10.73 & \cellcolor{gray!20}14.30 & 13.79 & 10.86 & 14.80 & \cellcolor{gray!20}16.25 & 14.40 & 12.78 & 12.60 & \cellcolor{gray!20}14.96 & 4.552 & 1.687 & 10 & \cellcolor{gray!20}10 \\
FlowEdit (SD3) \cite{Flowedit} & 12.07 &\cellcolor{gray!20}15.38 & 12.56 & 11.22 & 13.04 & \cellcolor{gray!20}14.82 & 12.86 & 11.45 & 11.43 & \cellcolor{gray!20}15.20 & 14.46 & 13.37 & 12.30 & \cellcolor{gray!20}14.85 & 4.542 & 1.676 & 11 & \cellcolor{gray!20}11 \\
Bagel \cite{bagel} & 13.33 & \cellcolor{gray!20}15.67 & 13.23 & 11.30 & 10.46 & \cellcolor{gray!20}14.26 & 13.61 & 11.83 & 13.41 & \cellcolor{gray!20}15.76 & 15.46 & 8.310 & 12.16 & \cellcolor{gray!20}14.86 & 4.040 & 1.670 & 12& \cellcolor{gray!20}12 \\
RFSE \cite{RFSE} & 12.44 & \cellcolor{gray!20}15.62 & 12.91 & 11.11 & 12.86 &\cellcolor{gray!20}15.11 & 14.63 & 11.40 & 10.81 & \cellcolor{gray!20}14.10 & 17.27 & 11.58 & 12.15 & \cellcolor{gray!20}14.71 & 4.100 & 1.678 & 13 & \cellcolor{gray!20}14\\
FlowEdit (FLUX) \cite{Flowedit} & 12.21 & \cellcolor{gray!20}15.46 & 12.94 & 10.70 & 12.26 & \cellcolor{gray!20}15.01 & 14.33 & 10.99 & 11.63 & \cellcolor{gray!20}14.50 & 13.97 & 12.17 & 12.09 & \cellcolor{gray!20}14.74 & 4.461 & 1.689 & 14& \cellcolor{gray!20}13 \\
ZONE \cite{zone} & 12.17 & \cellcolor{gray!20}15.19 & 12.54 & 10.95 & 9.934 &\cellcolor{gray!20}14.17 & 13.28 & 11.59 & 14.30 & \cellcolor{gray!20}16.21 & 12.42 & 12.62 & 11.84 & \cellcolor{gray!20}14.81 & 4.472 & 1.678 & 15 & \cellcolor{gray!20}15 \\
OT-Inversion \cite{OT} & 12.48 & \cellcolor{gray!20}15.62 & 12.44 & 12.26 & 12.14 & \cellcolor{gray!20}14.91 & 14.32 & 12.58 & 10.39 & \cellcolor{gray!20}14.06 & 16.54 & 10.47 & 11.74 & \cellcolor{gray!20}14.61 & 3.948 & 1.648 & 16 & \cellcolor{gray!20}16 \\
Any2Pix \cite{instructany2pix} & 9.270 & \cellcolor{gray!20}14.12 & 10.06 & 14.93 & 11.64 & \cellcolor{gray!20}14.41 & 13.89 & 14.79 & 8.384 & \cellcolor{gray!20}12.49 & 17.32 & 8.400 & 9.951 & \cellcolor{gray!20}13.51 & 3.459 & 1.583 & 17 & \cellcolor{gray!20}17 \\
MagicBrush \cite{Magicbrush} & 11.69 & \cellcolor{gray!20}14.02 & 12.36 & 11.41 & 8.504 & \cellcolor{gray!20}10.77 & 10.31 & 10.75 & 9.693 & \cellcolor{gray!20}13.10 & 12.22 & 12.34 & 9.772 & \cellcolor{gray!20}12.16 & 4.388 & 1.682 & 18 & \cellcolor{gray!20}18 \\
ACE++ \cite{ACE} & 8.925 & \cellcolor{gray!20}13.52 & 9.703 & 13.86 & 9.781 & \cellcolor{gray!20}11.86 & 12.67 & 13.76 & 8.905 & \cellcolor{gray!20}9.961 & 11.79 & 8.760 & 9.278 & \cellcolor{gray!20}11.54 & 3.832 & 1.610 & 19 & \cellcolor{gray!20}19 \\
IP2P \cite{ip2p} & 8.557 & \cellcolor{gray!20}12.93 & 9.904 & 11.33 & 8.637 & \cellcolor{gray!20}10.10 & 9.751 & 11.15 & 6.968 & \cellcolor{gray!20}9.844 & 11.38 & 13.35 & 8.116 & \cellcolor{gray!20}10.63 & 4.499 & 1.678 & 20 & \cellcolor{gray!20}20 \\
HQEdit \cite{HQ} & 8.834 & \cellcolor{gray!20}14.32 & 10.05 & 20.20 & 6.175 & \cellcolor{gray!20}8.588 & 8.420 & 20.20 & 5.275 & \cellcolor{gray!20}9.011 & 7.398 & 4.040 & 6.641 & \cellcolor{gray!20}10.08 & 2.655 & 1.441 & 21 & \cellcolor{gray!20}21 \\
PnP \cite{PnP} & 6.608 & \cellcolor{gray!20}12.23 & 8.570 & 11.90 & 5.917 & \cellcolor{gray!20}9.502 & 9.171 & 11.21 & 5.416 & \cellcolor{gray!20}9.431 & 9.591 & 12.97 & 5.959 & \cellcolor{gray!20}10.07 & 4.523 & 1.669 & 22 & \cellcolor{gray!20}22 \\
ReNoise \cite{renoise} & 1.544 & \cellcolor{gray!20}1.402 & 3.446 & 12.52 & 2.367 & \cellcolor{gray!20}2.193 & 2.400 & 11.19 & 1.719 & \cellcolor{gray!20}1.572 & 1.377 & 12.68 & 1.892 & \cellcolor{gray!20}1.736 & 4.361 & 1.661 & 23 & \cellcolor{gray!20}23 \\

\hline
\noalign{\vspace{1.5pt}}
SRCC to human $\uparrow$&&\cellcolor{gray!20}\mredbf{0.976}&
\mbluebf{0.951}&0.766&&\cellcolor{gray!20}\mredbf{0.987}&\mbluebf{0.952}&0.632&&\cellcolor{gray!20}\mredbf{0.996}&\mbluebf{0.734}&0.612&&\cellcolor{gray!20}\mredbf{0.998}&0.755& \mbluebf{0.833}&&\cellcolor{gray!20}\mredbf{0.998}\\
\bottomrule
\end{tabular}
}
\label{compare_TIE}
\end{table*}

\subsection{Ablation Study}
We conduct comprehensive ablation studies, with the results shown in Table~\ref{ablation}. Rows 1-2 illustrate the importance of the MLP head for precise score regression. Rows 2-4 demonstrate that combining pointwise and pairwise training yields the best performance. Rows 4-8 compare different MLLM backbones with similar parameter scales, among which InternVL3.5 achieves the best results.

\begin{table*}[tb]
\centering
\caption{Ablation study on the different backbones and training strategy.}
\label{ablation}
\setlength{\tabcolsep}{5pt} 
 \resizebox{1\textwidth}{!}{
\begin{tabular}{lccc|ccc|ccc|ccc}
\toprule
 \noalign{\vspace{-1.3pt}}
\multicolumn{4}{c}{Backbone$\&$Strategy} & \multicolumn{3}{c}{Perceptual Quality} & \multicolumn{3}{c}{Editing Alignment} & \multicolumn{3}{c}{Attribute Preservation}\\
 \noalign{\vspace{-1.0pt}}
\cmidrule(lr){1-4} \cmidrule(lr){5-7} \cmidrule(lr){8-10} \cmidrule(lr){11-13}
 \noalign{\vspace{-1.5pt}}
Backbone&MLP Head &  Pointwise Training & Pairwise Training&  $\mathrm{SRCC}_{global}$ & $\mathrm{SRCC}_{group}$ &Acc & $\mathrm{SRCC}_{global}$ & $\mathrm{SRCC}_{group}$ &Acc & $\mathrm{SRCC}_{global}$ & $\mathrm{SRCC}_{group}$ &Acc\\ 
\hline
\noalign{\vspace{1pt}}
InternVL3.5 \cite{internvl3_5} & & \checkmark &   & 0.7184 & 0.7215 &0.7904 & 0.7025&0.7312 &0.7750 & 0.7241&0.7506 &0.7862\\
InternVL3.5 \cite{internvl3_5}&\checkmark & \checkmark &  & 0.7762 & 0.7829 &0.7826 & 0.7806&0.7964 &0.7825 & 0.7752&0.7864 &0.7821\\
InternVL3.5 \cite{internvl3_5}&\checkmark &  & \checkmark  & 0.8018 & 0.8792 &0.8663 & 0.8213&0.8894 &0.8762 & 0.8173&0.8726 &0.8765\\
\rowcolor{gray!20}  
InternVL3.5 \cite{internvl3_5}&\checkmark & \checkmark &  \checkmark  & \textbf{0.8548} & \textbf{0.9223} &\textbf{0.9152} & \textbf{0.8488}&\textbf{0.9319} &\textbf{0.9298} & \textbf{0.8475}&\textbf{0.9251} &\textbf{0.9197}\\
InternVL3 \cite{internvl2}&\checkmark & \checkmark &  \checkmark  & 0.8316 & 0.8964 &0.8826 & 0.8155&0.9109 &0.9035 & 0.8141&0.9022 &0.8943\\
Qwen3-VL \cite{qwen3}&\checkmark & \checkmark &  \checkmark & 0.8417 & 0.9032 &0.8985 & 0.8231&0.9127 &0.9046 & 0.8240&0.9068 &0.9074\\
DeepSeekVL2 \cite{deepseekv2} &\checkmark & \checkmark &  \checkmark  & 0.8274 & 0.8823 &0.8970 & 0.8285&0.8963&0.8856&0.8106&0.8769&0.8811\\
LLaVA-NeXT \cite{llava-n}&\checkmark & \checkmark &  \checkmark  & 0.8164 & 0.8723 &0.8657 & 0.8108&0.8814 &0.8741 & 0.8037&0.8806 &0.8794\\

 \noalign{\vspace{-1.5pt}}
\bottomrule
\end{tabular}}
\end{table*}

\subsection{Zero-shot Performance on Other Benchmarks}
Table~\ref{other_dataset} presents the performance of EditHF compared with other advanced closed-source and open-source MLLMs on several image editing benchmarks. For benchmarks with pairwise annotations, including GenAI-Bench~\cite{genai}, AROURA-Bench~\cite{aurora}, and EditScore-Bench~\cite{editscore}, we report prediction accuracy. For benchmarks with pointwise annotations, including ImagenHub~\cite{imagenhub}, EBench-18K~\cite{lmm4edit}, and EditReward-Bench~\cite{editreward}, we report SRCC with human ratings. For benchmarks that provide multiple evaluation dimensions, including ImagenHub~\cite{imagenhub}, EBench-18K~\cite{lmm4edit}, EditScore-Bench~\cite{editscore}, and EditReward-Bench~\cite{editreward}, we report the average result across different dimensions. The results show that EditHF consistently achieves strong performance across diverse benchmarks, demonstrating its robust generalization capability.
\begin{table*}[tb]
\centering
\caption{Comparison results on other public image editing benchmarks. The best results are highlighted in \mredbf{red}, and the second-best results are highlighted in \mbluebf{blue}. “-” indicates that the model was trained on this dataset, and thus its result is not reported to ensure a fair cross-dataset evaluation.}
\label{other_dataset}
\setlength{\tabcolsep}{8pt} 
 \resizebox{1\textwidth}{!}{
\begin{tabular}{l||cccccc}
\toprule
 \noalign{\vspace{-1.3pt}}
Methods/Benchmarks&GenAI-Bench \cite{genai}&AROURA-Bench \cite{aurora} &ImagenHub \cite{imagenhub} &EBench-18K \cite{lmm4edit}&EditScore-Bench \cite{editscore}& EditReward-Bench \cite{editreward}\\
\hline
GPT-4o \cite{chatgpt4o}&0.5354&0.5081&0.3821&0.4108&0.6730&0.2831\\
GPT-5 \cite{gpt5}&0.5961&0.4727&0.4085&0.4260&\mbluebf{0.7524}&0.3781\\
Gemini-2.0-Flash \cite{gemini2}&0.5332&0.4431&0.2369&0.3892&0.6958&0.3347\\
Gemini-2.5-Flash \cite{nanobanana}&0.5701&0.4763&\mbluebf{0.4162}&\mbluebf{0.4337}&0.7026&0.3802\\
Qwen3-VL (7B) \cite{qwen3}&0.4458&0.3428&0.2426&0.3726&0.4251&0.2731\\
InternVL3.5 (7B) \cite{internvl3_5}&0.4517&0.3692&0.2071&0.3511&0.4518&0.2406\\
EditScore (Qwen3) \cite{editscore}&0.5024&0.4218&0.3062&0.3206&-&0.2590\\
EditReward (MiMo) \cite{editreward}&\mbluebf{0.6572}&0.6362&0.3520&0.3718&0.7125&-\\
LMM4Edit \cite{lmm4edit}&0.6327&\mbluebf{0.6521}&0.3726&-&0.7064&\mbluebf{0.3958}\\
\hline
\noalign{\vspace{1pt}}
\rowcolor{gray!20}  
EditHF (Ours)&\mredbf{0.6792}&\mredbf{0.6608}&\mredbf{0.4526}&\mredbf{0.4627}&\mredbf{0.7750}&\mredbf{0.4325}\\
 \noalign{\vspace{-1.5pt}}
\bottomrule
\end{tabular}}
\end{table*}
\setlength{\tabcolsep}{3pt} 
\begin{table*}[t]
\centering
\caption{Results of Qwen-Image-Edit \cite{qwenedit} refined with our EditHF on different benchmarks. Overall scores from human study and our EditHF are reported.}
\label{DPO}
\setlength{\tabcolsep}{4pt} 
 \resizebox{1\textwidth}{!}{
\begin{tabular}{l||cccccccccccc}
\toprule
 \noalign{\vspace{-1.3pt}}
Benchmarks&\multicolumn{2}{c}{EditHF-1M} &\multicolumn{2}{c}{AROURA-Bench \cite{aurora}} &\multicolumn{2}{c}{ImagenHub \cite{imagenhub}} &\multicolumn{2}{c}{EBench-18K \cite{lmm4edit}}&\multicolumn{2}{c}{EditScore-Bench \cite{editscore}}&\multicolumn{2}{c}{EditReward-Bench \cite{editreward}}\\
\cmidrule(lr){2-3}
\cmidrule(lr){4-5}
\cmidrule(lr){6-7}
\cmidrule(lr){8-9}
\cmidrule(lr){10-11}
\cmidrule(lr){12-13}
 \noalign{\vspace{-1.3pt}}
Methods/Metrics&Human&Ours&Human&Ours&Human&Ours&Human&Ours&Human&Ours&Human&Ours\\
\hline
Qwen-Image-Edit&75.39&78.06&83.21&68.47&75.64&73.26&73.25&78.29&70.17&75.22&80.15&77.03\\
\hdashline
\noalign{\vspace{1pt}}
Qwen-Image-Edit (Self-DPO)&78.36&81.55&84.28&69.26&76.55&75.14&73.80&79.05&72.18&76.55&81.17&78.46\\
\rowcolor{gray!20}  
Improvement& \mredbf{+3.94\%} & \mredbf{+4.47\%} & \mredbf{+1.29\%} & \mredbf{+1.15\%} & \mredbf{+1.20\%} & \mredbf{+2.57\%} & \mredbf{+0.75\%} & \mredbf{+0.97\%} & \mredbf{+2.86\%} & \mredbf{+1.77\%} & \mredbf{+1.27\%} & \mredbf{+1.86\%} \\
\hdashline
\noalign{\vspace{1pt}}
Qwen-Image-Edit (Global-DPO)&84.26&86.15&85.46&69.12&79.12&78.06&74.18&79.60&75.23&78.41&82.60&78.12\\
\rowcolor{gray!20}  
Improvement& \mredbf{+11.77\%} & \mredbf{+10.36\%} & \mredbf{+2.70\%} & \mredbf{+0.95\%} & \mredbf{+4.60\%} & \mredbf{+6.55\%} & \mredbf{+1.27\%} & \mredbf{+1.67\%} & \mredbf{+7.21\%} & \mredbf{+4.24\%} & \mredbf{+3.06\%} & \mredbf{+1.42\%} \\
 \noalign{\vspace{-1.5pt}}
\bottomrule
\end{tabular}}
\end{table*}
\subsection{EditHF for Reward Modeling}
To validate the effectiveness of EditHF as a reward model for refining image editing models, we apply it to refine Qwen-Image-Edit~\cite{qwenedit} that achieves the best performance on EditHF-1M among all open-source editing models. 

We first select source images and their corresponding instruction prompts from the EditHF-1M training set across all editing tasks. For each instruction, we use Qwen-Image-Edit to generate edited images with 20 different random seeds, and subsequently evaluate all generated results using EditHF. Within each seed group, we designate the sample with the highest average score across all evaluation dimensions as the chosen sample and the lowest-ranked one as the rejected sample. To avoid constructing preferences from excessively ambiguous or failure-prone cases, we exclude instructions whose edits consistently receive low overall scores. Moreover, within each editing task, we further rank the resulting preference pairs according to the overall score of the chosen sample and discard the lower half, ensuring that only relatively high-quality edits are retained for preference construction. In total, we obtain 18K high-quality preference pairs.

We then employ Direct Preference Optimization (DPO) to align the outputs of Qwen-Image-Edit with human preferences. Specifically, we freeze the VAE module and optimize the Transformer by comparing the flow prediction errors between the original pre-trained Qwen-Image-Edit, which serves as the reference model, as follows:
\begin{equation}
\begin{aligned}
\mathcal{L}(\theta)
= - \mathbb{E}_{(x_0^c, x_0^r) \sim \mathcal{D}_{\mathrm{Gen}},\, t \sim \mathcal{U}(0, T)}
\Big[
\log \sigma \big(
\Delta_\theta(x_t^c, t) - \Delta_\theta(x_t^r, t)
\big)
\Big], \\
\Delta_\theta(x_t, t)
= \beta_g \Big(
\lVert (\epsilon - x_0) - v_\theta(x_t, t) \rVert_2^2
- \lVert (\epsilon - x_0) - v_{\mathrm{ref}}(x_t, t) \rVert_2^2
\Big).
\end{aligned}
\end{equation}
where $\epsilon$ denotes the Gaussian noise sampled from a standard normal distribution, which is used to perturb the clean image $x_0$. $x_t^c$ and $x_t^r$ denote the noisy latents of the chosen and rejected samples at timestep $t$, respectively.
$v_\theta(\cdot)$ and $v_{\mathrm{ref}}(\cdot)$ denote the flow predictions of the fine-tuned and reference editing models. $\beta_g$ is a temperature parameter controlling the optimization strength,
and $\sigma(\cdot)$ denotes the logistic function. This loss encourages the fine-tuned editing model to reduce the denoising error for chosen samples while increasing it for rejected ones, while referencing the editing model before fine-tuning to enhance training stability.

Relying on a model to generate its own reward signals may reinforce its inherent biases and overlook alternative high-quality outputs. To obtain a more diverse and reliable supervision signal, we construct chosen and rejected samples using outputs from multiple editing models in EditHF-1M. Specifically, for each image group containing edited results produced by different models under the same source image and instruction, we designate the highest-ranked image as the chosen sample and the lowest-ranked one as the rejected sample. To avoid excessively challenging or ambiguous cases, we exclude instructions with low overall scores across all editing models. Following this process, we construct 22K preference pairs and fine-tune Qwen-Image-Edit using the same training strategy described above.

\begin{figure}[t]
    \centering
    \includegraphics[width=1\linewidth]{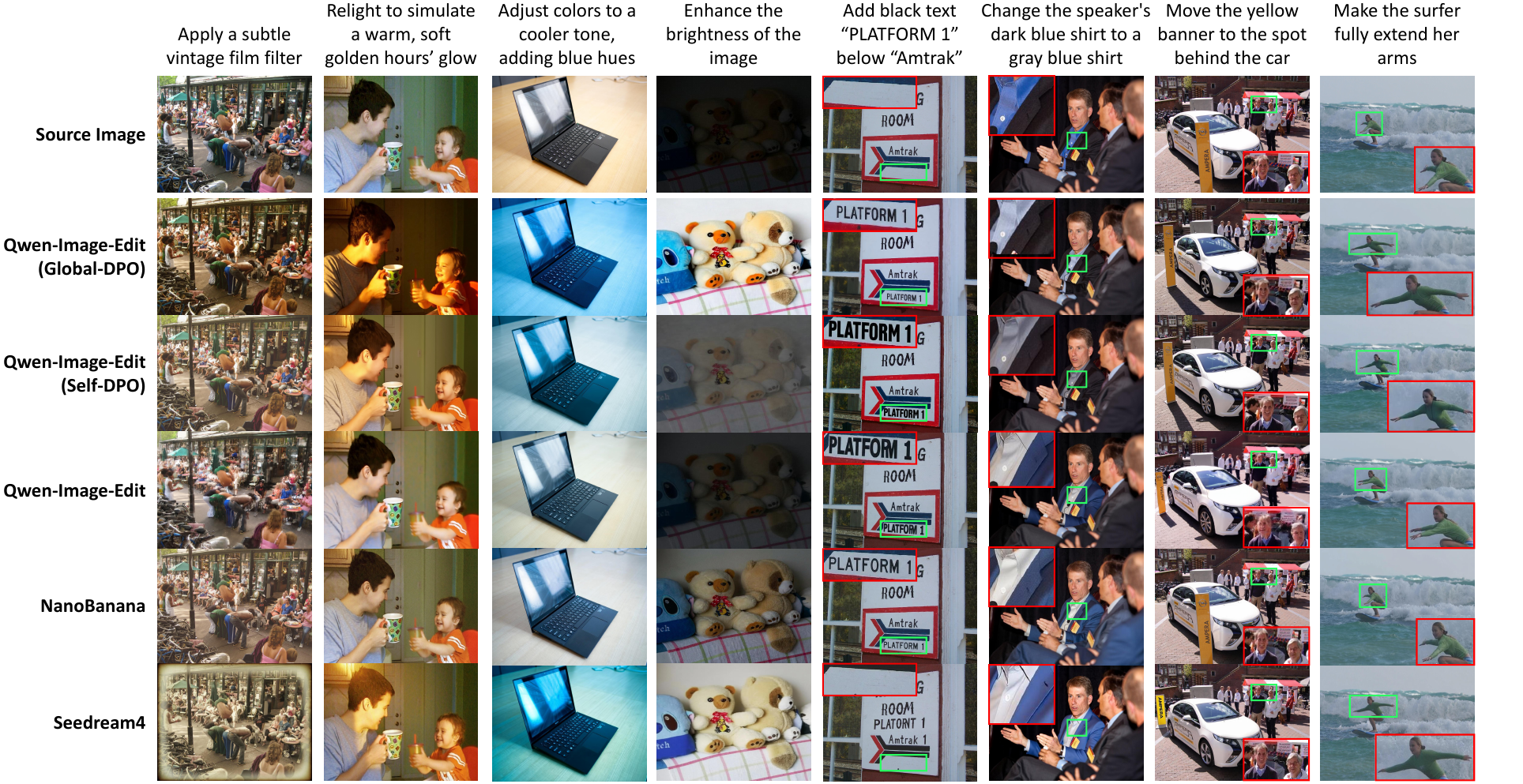}
    \caption{Examples from Qwen-Image-Edit refined by our EditHF and other competitive editing models.}
    \label{visualization}
\end{figure}

We then compare the original Qwen-Image-Edit with versions refined using its own outputs (Self-DPO) and using outputs from multiple models (Global-DPO) across several image editing benchmarks. We conduct a subjective study following the same protocol of scoring annotation described in Section~\ref{3.3}. The results in Table~\ref{DPO} show that Qwen-Image-Edit refined using our EditHF consistently achieves performance gains across these benchmarks in both human evaluations and the EditHF metric, with greater gains observed when using Global-DPO. Figure~\ref{visualization} presents representative editing examples produced by refined Qwen-Image-Edit, alongside those from other competitive editing models, demonstrating that our refined model exhibits a strong capability to handle fine-grained editing tasks with higher fidelity and better alignment with the given instructions.
\section{Conclusion}
In this paper, we introduce EditHF-1M, the first million-scale benchmark for multi-dimensional evaluation of image editing, including both preference pairs and fine-grained scores from rigorous human annotations. Based on this benchmark, we propose EditHF, which effectively supports both pointwise scoring and pairwise comparison. Extensive experiments show that EditHF outperforms existing methods, achieving stronger alignment with human preferences and better generalization. Moreover, we demonstrate that EditHF can serve as an effective reward model for refining image editing models.

%
%
\clearpage
\bibliographystyle{splncs04}
\bibliography{main}
\end{document}